# Spatiotemporal Prediction of Secondary Crashes by Rebalancing Dynamic and Static Data with Generative Adversarial Networks

Junlan Chen[1, 2], Yiqun Li[1], Chenyu Ling [1], Ziyuan Pu[1], Xiucheng Guo[1]

[1]School of Transportation, Southeast University, No.2 Southeast University Road, Nanjing, China, 211189. E-mail: junlan.chen@monash.edu; lyq.yiqun@foxmail.com; lingchenyu1999@126.com; ziyuan.pu@monash.edu; seuguo@163.com

[2]School of Civil Engineering, Monash University, Melbourne, Australia. E-mail: junlan.chen@monash.edu

**ABSTRACT:**

Data imbalance is a common issue in analyzing and predicting sudden traffic events. Secondary crashes constitute only a small proportion of all crashes. These secondary crashes, triggered by primary crashes, significantly exacerbate traffic congestion and increase the severity of incidents. However, the severe imbalance of secondary crash data poses significant challenges for prediction models, affecting their generalization ability and prediction accuracy. Existing methods fail to fully address the complexity of traffic crash data, particularly the coexistence of dynamic and static features, and often struggle to effectively handle data samples of varying lengths. Furthermore, most current studies predict the occurrence probability and spatiotemporal distribution of secondary crashes separately, lacking an integrated solution. To address these challenges, this study proposes a hybrid model named VarFusiGAN-Transformer, aimed at improving the fidelity of secondary crash data generation and jointly predicting the occurrence and spatiotemporal distribution of secondary crashes. The VarFusiGAN-Transformer model employs Long Short-Term Memory (LSTM) networks to enhance the generation of multivariate long-time series data, incorporating a static data generator and an auxiliary discriminator to model the joint distribution of dynamic and static features. In addition, the model's prediction module achieves simultaneous prediction of both the occurrence and spatiotemporal distribution of secondary crashes. Compared to existing methods, the proposed model demonstrates superior performance in generating high-fidelity data and improving prediction accuracy.

## 1 INTRODUCTION

Data imbalance is a common issue in many fields, such as traffic management and medical diagnosis, posing significant challenges for prediction and analysis. In the field of traffic management, secondary crashes are typically defined as those triggered by a primary crash within a certain spatial and temporal range. These secondary crashes can exacerbate traffic congestion, increase the severity of incidents, and even lead to more casualties. Thus, preventing secondary crashes has become an important task in traffic safety management. However, due to the inherent data imbalance (secondary crashes usually account for less than 2% of all crashes), predicting secondary crashes remains highly challenging, as the imbalance significantly affects the generalization ability of prediction models. Moreover, existing studies tend to focus on minimizing overall errors in temporal and spatial predictions, often overlooking the precision in the number of secondary crashes. Additionally, the limited data available for secondary crashes makes it difficult for prediction models to capture sufficient feature patterns. Therefore, addressing the imbalance in secondary crash data is of utmost importance.

Researchers have proposed various methods to address data imbalance, which can be broadly categorized into dynamic data generation methods and static data generation methods. Dynamic data generation methods primarily target continuous time-series data. These methods often employ deep learning-based generative models, such as Time Series Generative Adversarial Networks (TimeGAN) (7) and Variational Autoencoders (VAE) (8; 9), to generate new time-series data. However, for multivariate, long time-series traffic crash data, existing generative models face challenges in capturing the complex temporal and spatial relationships within the data. Compared to short time-series data, the dependencies between data points in long time series become more complex as the time span increases, undergoing multiple nonlinear transformations, which makes it difficult for models to effectively capture them (10). Additionally, as time progresses, more variables and interfering factors further increase the learning difficulty (11). Static data generation methods mainly target discrete data or independent continuous data unaffected by time. These methods include traditional over-sampling and under-sampling techniques, such as Synthetic Minority Over-sampling Technique (SMOTE) (12) and Adaptive Synthetic Sampling (ADASYN) (13). These methods balance datasets by generating new samples or adjusting the weights of existing samples. Due to the complex feature distribution of traffic crash data, simple over-sampling or under-sampling methods often fail to effectively capture the intrinsic patterns of the data, leading to suboptimal performance when dealing with complex traffic crash data (14). Furthermore, as secondary crashes may occur at different time points after a primary crash, the sample data for secondary crashes contains complex multidimensional data of varying time lengths, which existing generative models cannot effectively handle. Overall, current generative models face three main issues when handling traffic crash data: First, for multivariate, long-time series data, generative models perform poorly and struggle to capture the complex temporal and spatial relationships within the data. Second, no suitable model exists to generate complex multidimensional traffic crash data combining both static and dynamic data. Third, existing methods cannot generate samples of varying lengths. Therefore, effectively generating high-fidelity secondary crash data remains a significant challenge.

Current research on secondary crashes can be divided into identification and prediction. In terms of identification, early studies mainly used fixed spatiotemporal threshold methods, assuming secondary crashes occur within a certain spatiotemporal range of the initial crash. The limitation of this method lies in the subjective determination of thresholds. To overcome this limitation, subsequent research proposed various dynamic identification methods, such as speed contour methods. In terms of prediction, existing studies mainly explore the impact of initial crash characteristics, weather conditions, and traffic volume on the likelihood of secondary crashes. In recent years, with the rapid development of machine learning technologies, algorithms such as Support Vector Machines and Neural Networks have been widely applied to secondary crash prediction, achieving better predictive performance than traditional Logit models. Despite this, most existing studies separate the likelihood of secondary crash occurrence from their spatiotemporal distribution, predicting them independently. In fact, predicting the specific spatiotemporal distribution is only meaningful under the premise that a crash has indeed occurred. Therefore, constructing a dynamic model capable of simultaneously predicting the probability and spatiotemporal distribution of secondary crashes is crucial for uncovering the intrinsic relationships between crashes and developing proactive crash management strategies.

In summary, current research on secondary crash prediction faces three main problems: first, the severe imbalance of secondary crash data affects the generalization ability and accuracy of predictive models. Second, existing methods for handling imbalanced data are insufficient to address the complexity of traffic crash data, where dynamic and static features coexist, particularly the inability to generate samples of varying lengths. Lastly, current studies often predict the probability of secondary crash occurrence and their spatiotemporal distribution separately, lacking a dynamic model that can predict both simultaneously. In light of these issues, this paper proposes a hybrid model named **VarFusiGAN-Transformer**, designed to enhance the fidelity of secondary crash data generation and jointly predict the occurrence probability and spatiotemporal location of secondary crashes. The specific contributions of this paper are as follows:

1. A Long Short-Term Memory (LSTM) network is used instead of Multilayer Perceptrons (MLP) to develop a new generator in the DG. Batch generation methods are used to capture complex dependencies between variables, improving the fidelity of dynamic data generation for multivariate, long-time series.

2. A static data generator and an auxiliary discriminator are added to the generative model to simulate the joint distribution of static and dynamic data, achieving the generation of complex traffic crash data.

3. A machine learning model named VarFusiGAN-Transformers is constructed to simultaneously predict the occurrence probability and spatiotemporal distribution of secondary crashes.

## 2 LITERATURE REVIEW

With the rapid development of deep learning technology, dynamic data generation methods have made significant progress in handling time series data. These generation methods primarily target continuous time series data by capturing the intrinsic spatiotemporal relationships within the data to achieve data augmentation. The GAN model has been widely applied in time series generation due to its flexibility and specialized generation mechanism. Yoon et al. proposed a generation framework called TimeGAN, which leverages the flexibility of GAN and the control capability of autoregressive models over temporal dynamics. It uses an embedding network to map data into a low-dimensional latent space to enhance training effectiveness, generating high-quality time series samples across multiple datasets(*7*). For imbalanced crash data, Cai et al. introduced convolutional neural networks to construct generators and discriminators to better recognize and capture traffic data features(*8*). Additionally, models such as SigCWGAN(*25*), TSGAN(*26*) and TTS-CGAN(*27*) have also achieved time series data generation. However, the aforementioned methods still cannot resolve issues like mode collapse caused by GAN models. Therefore, researchers have attempted other methods in recent years. For instance, Desai et al. combined variational autoencoders (VAE) with time series characteristics and proposed a generation model named TimeVAE(*9*). Lee et al. used vector quantization to map time series into a discrete latent space and employed a bidirectional Transformer model for dynamic capturing, achieving significant performance improvements in time series generation tasks(*27*). However, due to the complex spatio-temporal dependencies in long time series traffic data and the introduction of disturbing variables over time, existing generative models still face many challenges in improving the fidelity of the generated data.

It is worth noting that generative models such as VAE and GAN are mainly designed for continuous variables, while in traffic crash data, dynamic data and static data often exist at the same time. Therefore, how to effectively generate composite samples containing these two types of data is an important and challenging research topic. Traditional oversampling(*28*) and undersampling(*29*) techniques have many limitations when processing such complex data. These methods may destroy the intrinsic structure of time series data, resulting in information loss, and may generate synthetic samples that do not conform to the actual situation(*30*). To address this, some researchers have developed advanced generative models suitable for both dynamic and static variables. For instance, Chen et al. proposed a hybrid resampling method based on Conditional Tabular GAN (CTGAN-RU) to tackle the imbalance problem of fatal crashes relative to non-fatal crashes. This method effectively handles scarce discrete variables and generates high-fidelity composite data(*31*). Park et al. combined deep convolutional GAN with classifier neural networks to propose a model called TableGAN(*32*), which improves the consistency of two types of data in generated records. Medical Wasserstein GAN (MedWGAN) uses gradient penalty and boundary-seeking GAN to generate more realistic synthetic medical records(*33*). However, the high-dimensional and long time series traffic crash data exhibit complex dynamic relationships and mixed distribution characteristics(*10*). The data sparsity and long-term memory challenges increase the difficulty of the generation task, making it difficult for existing methods to generate data that meets the requirements of crash prediction models.

Since traffic crash data usually does not categorize primary and secondary crashes, identifying secondary crashes is typically the first step in research. Early methods for identifying secondary crasehes mainly used static approaches. These methods utilized predefined spatiotemporal thresholds to identify secondary crashes. For example, Raub(*15*) first proposed using spatiotemporal thresholds of 15 minutes and 1 mile to identify secondary crashes. Subsequently, Moore(*16*) and Hirunyanitiwattana(*17*) selected different ranges for secondary crash identification. However, the fixed thresholds of static methods cannot accurately reflect changes in actual traffic conditions, leading to the application of dynamic methods in secondary crash identification(*18*). Dynamic methods include approaches based on queue theory(*34*), shockwave theory(*35*), and speed contour maps(*18*). Among these, the speed contour map method determines the spatiotemporal impact range of crashes using real-time traffic flow data and is currently one of the more popular methods in research. Secondary crash prediction methods mainly focus on predicting the likelihood of a primary crash causing a secondary crash and the spatiotemporal distribution of secondary crashes. Early prediction models often used statistical models, with the logit model being one of the most commonly used methods(*36; 37*). In recent years, with the development of machine learning and deep learning technologies, prediction models based on these technologies have gradually become mainstream. For example, Park et al. proposed a secondary crash prediction model based on Bayesian Neural Network (BNN), achieving better results than traditional Logit models(*24*). Liu et al. used an improved Random Forest model to predict the spatiotemporal location of secondary crashes, showing that this model outperforms K-Nearest Neighbors and Multilayer Perceptron Regression models in prediction performance(*4*). Additionally, hybrid models combining multiple deep learning algorithms can be considered for predicting the spatiotemporal location of secondary crashes. More complex models may have better predictive performance. For instance, Li et al. proposed a hybrid deep learning model that combines Stacked Sparse Autoencoders (SSAE) and Long Short-Term Memory (LSTM) networks, capable of extracting key features, capturing long-term dependencies, and performing nonlinear fitting. The results indicate that compared to single models, the hybrid model performs better in both spatial and temporal predictions(*3*).

## 3 METHODS

### 3.1 Overview of VarFusiGAN-Transformer Model

This study proposes a hybrid model named VarFusiGAN-Transformer, which aims to predict both the occurrence probability and the spatiotemporal distribution of secondary crashes by integrating generative and predictive components. The model consists of two main parts: the generative component (VarFusiGAN) and the predictive component (Transformer). The generative component is responsible for generating high-fidelity synthetic secondary crash data, while the predictive component performs precise spatiotemporal predictions. By combining these components, the VarFusiGAN-Transformer model enhances data diversity, improves data balance, and ultimately increases prediction accuracy and robustness for secondary crashes. The structure of the model is shown in **Figure 3**.

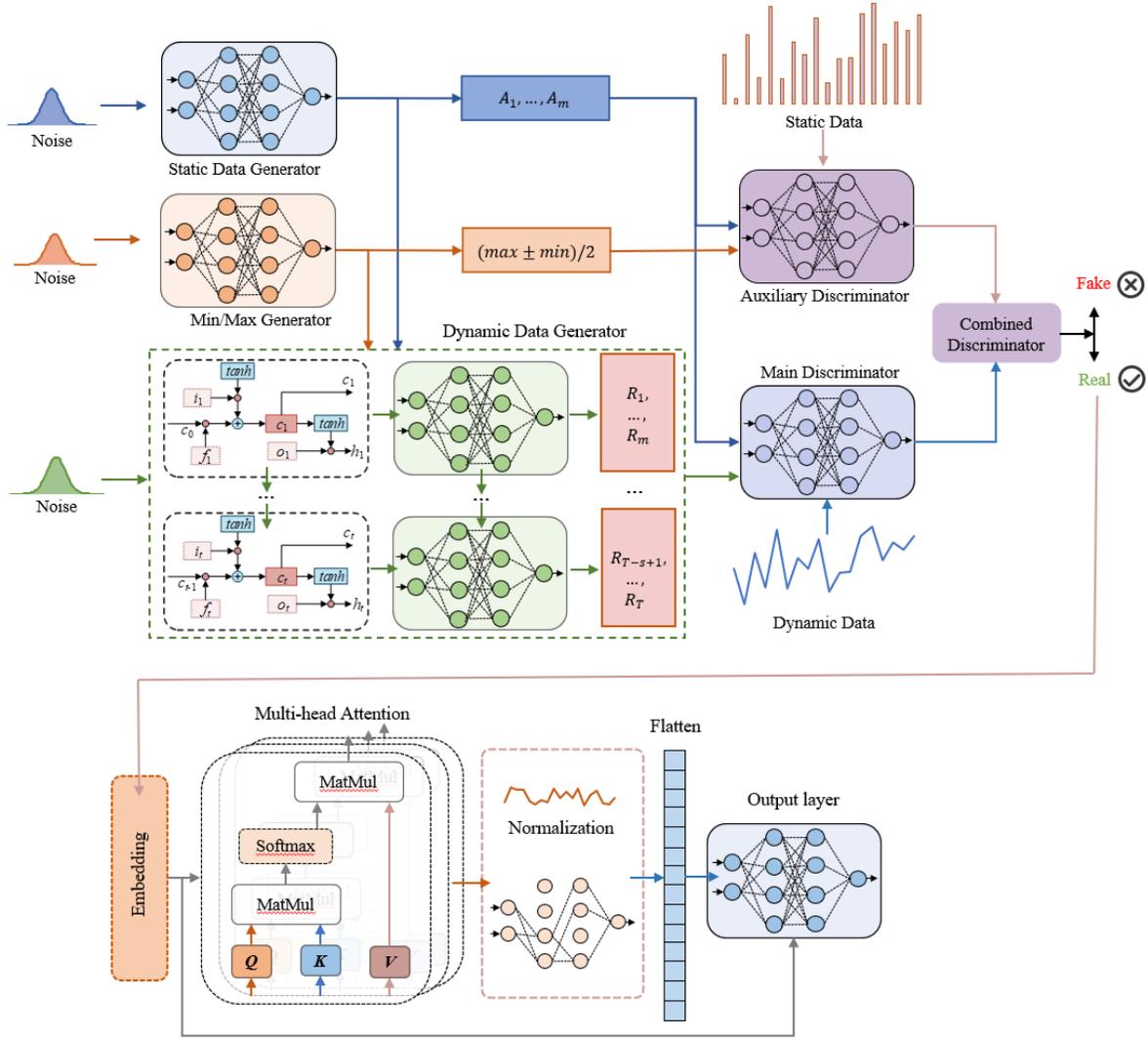

**Figure 3 The Structure of VarFusiGAN-Transformer Model**

*3.1.1 Generative Component: VarFusiGAN*

The generative component, VarFusiGAN, is designed to address the challenges of generating multidimensional composite data that include both dynamic (time-series) and static features. The VarFusiGAN model leverages Generative Adversarial Networks (GANs) and incorporates the following key innovations:

**(1) Capturing Multivariable Correlations**

In the research related to traffic crashes, data samples usually contain more variables and longer time series. This imposes higher demands on the generator to capture the multivariable correlations in dynamic data. Traditional GAN generators typically use a fully connected multilayer perceptron (MLP) architecture, which performs poorly in capturing complex inter-variable relationships. Long Short-Term Memory (LSTM) networks, designed for modeling time series, have been widely used in GAN models for generating time series data. Therefore, we use LSTM instead of MLP for generating multivariable time series data.

For a single variable, unlike MLPs which generate an entire time series at once, LSTMs generate one record $R_j^i$ at a time (the variable $i$ at the $j$th time step) and run for T (the number of time steps) iterations to generate the entire time series for that variable. The main feature of LSTM is its internal state, which can encode and store all past states. For example, when generating the flow value at the $j$th time step, it can combine all flow values from the previous $j-1$ time steps, ensuring the fidelity of a single variable over the entire period.

Due to the complex dependencies between variables, traditional LSTMs often face challenges when handling multiple variables. This is because LSTM needs to update and remember each variable at every time step. As the number of variables increases, the complexity of these updates and memories also increases, affecting the quality of the generated data. To address this, we propose a batch generation method to reduce the number of passes. Instead of generating a record for a single variable at each generation step, LSTM generates records for $S$ variables (e.g., the variable 1, 2 and 3 at $j$th time step). This can reduce the number of passes by a factor of $S$. However, it is important to note that as $S$ increases, although the number of passes decreases, the difficulty of each generation step increases, requiring multiple experiments for parameter tuning. In practice, we found that $S = 3$ is most effective for the secondary crash dataset in this study.

**(2) Addressing the Mode Collapse Problem**

GAN models often encounter mode collapse issues when handling multivariable traffic time series data due to the high variability in traffic flow characteristics. For example, at a certain time step, the traffic flow in many samples might generally exceed 100, while the flow differences between adjacent lanes are less than 10 or even 0. Traditional GAN models typically use standard normalization methods, scaling the data based on the maximum and minimum values of the entire dataset and training accordingly. This approach often leads to homogenized output samples, triggering mode collapse. To address this, VarFusiGAN introduces an automatic normalization mechanism. This method normalizes each dynamic variable individually and incorporates the maximum and minimum values of each variable as "pseudo" static data into the learning process. For instance, we normalize the data for the upstream average flow variable across all 6-time steps and record the maximum and minimum values within this range. During generation, these values are used to constrain the variable, thus alleviating the mode collapse problem. The two normalization methods described above can be specified as shown in

**Equation 1**

$$x_{norm} = \frac{x - x_{\min}}{x_{\max} - x_{\min}}$$
$$x^i_{j,norm} = \frac{x^i_j - x^i_{\min}}{x^i_{\max} - x^i_{\min}} \tag{1}$$

where $x_{\max}$ and $x_{\min}$ represent the minimum and maximum values in the entire dataset, $x_{norm}$ represents the normalized value of any variable $x$, and $x^i_{j,norm}$ denotes the normalized value of a specific variable $i$ at time step $j$.

## (3) Capturing Relationships between Static and Dynamic Data

In crash data, there is often a close relationship between static and dynamic data. For example, more severe crashes often occur when vehicle speeds are higher prior to the incident. Therefore, we need a mechanism to model the joint distribution between static and dynamic data.

Regarding the generator, VarFusiGAN's generator consists of two independent parts: a static data generator and a dynamic data generator. The static data generator takes a random noise vector as input and uses a multilayer perceptron (MLP) to generate the static data. This generated static data is then passed as input to the dynamic generator at each step. The dynamic generator (i.e., the LSTM) generates time series data. Through this decoupled generation strategy, VarFusiGAN can better capture the complex relationships between static and dynamic data in the samples.

Experiments have shown that when dealing with high-dimensional sample data, a single discriminator often struggles to ensure the fidelity of the generated data, especially the static data. To address this, VarFusiGAN introduces an auxiliary discriminator that works in conjunction with the primary discriminator, with their combined loss function shown in **Equation 2**. $L_1(G, D_1)$ is the loss function for the auxiliary discriminator. $D_1(a)$ and $D_1(G(z)$ represent the probabilities that real static data $a$ and generated static data $G_a(z)$ are judged as real, respectively. This discriminator is solely used to determine whether the generated static data is authentic. The primary discriminator $L_2(G, D_2)$, on the other hand, inputs real data $x$ that includes both static and dynamic features to determine whether the generated dynamic data matches the static data. Finally, the losses of the two discriminators are combined using a weighting parameter to form a joint discriminator, which can be optimized by adjusting the parameter $\alpha$ to enhance the overall generative performance.

$$L_1(G, D_1) = \mathbb{E}_{a \sim p_{data}(x)}[\log D_1(a)] + \mathbb{E}_{z \sim p_z(z)}[\log(1 - D_1(G_a(z)))]$$

$$L_2(G, D_2) = \mathbb{E}_{x \sim p_{data}(x)}[\log D_2(x)] + \mathbb{E}_{z \sim p_z(z)}[\log(1 - D_2(G(z)))] \qquad (2)$$

$$\min_G \max_{D_1, D_2} \alpha L_1(G, D_1) + L_2(G, D_2)$$

**Figure 3** illustrates the structure of the VarFusiGAN model. The auxiliary generator assesses the authenticity of static data such as crash type, severity, and spatiotemporal distance, while the primary generator evaluates whether the dynamic variables related to traffic flow characteristics before the crash (e.g., flow, occupancy, and speed) match the static variables.

### (4) Handling Sample Data of Different Length

In traffic crash analysis, secondary crashes are typically a series of events that occur shortly after a primary crash and may happen at different time points. Therefore, when collecting data samples, the length of dynamic data for each sample varies depending on the time of the secondary crash occurrence. For example, when a secondary crash occurs at the 6 minute after a primary crash, we need to collect data from 30 minutes before the primary crash to 5 minutes after the crash; whereas if the secondary crash occurs at the 11th minute, we need to collect data from 30 minutes before to 10 minutes after the primary crash. As a result, the dynamic data lengths of different samples are different. However, current generative models often struggle to manage datasets with different sample lengths effectively. Applying a straightforward truncation method to standardize sample lengths by removing excess data can lead to the omission of crucial temporal information, particularly in real-time prediction contexts where the discarded data might contain key traffic flow characteristics around the occurrence of the secondary crash, which would inevitably affect the prediction accuracy.

However, the VarFusiGAN model used in this paper proposes a unique solution to this problem, i.e., handling samples of different lengths by introducing the data_gen_flag field.

data_gen_flag is a floating-point array of size [number of training samples × maximum time length] used to mark the activation state of each sample at each time step. Where 1 indicates that the time step is active and 0 indicates that the time step is not active. This allows flexibility in representing time series data of different lengths. For example, suppose there are two samples with lengths of 2 and 4, and the maximum time length is set to 4. Then data_gen_flag will be displayed as:

data_gen_flag=[
[1.0,1.0,0.0,0.0]              (3)
[1.0,1.0,1.0,1.0]]

For the first sample of length 2, data_gen_flag is 1 at the 1st and 2nd time steps and 0 at the 3rd and 4th time steps, indicating that only the data in the first two time steps are valid; while for the second sample of length 4, data_gen_flag is 1 at all 4 time steps, indicating that the entire time series is valid. In this way, the model is able to dynamically adapt to samples of varying lengths during generation and prediction, ensuring that the generated data have high fidelity.

With the above setup, the VarFusiGAN model can generate datasets containing both dynamic features and static attributes, and is able to handle samples of different lengths, maintaining data consistency and integrity. In this way, the model can generate high-fidelity data while ensuring the diversity and accuracy of the generated data, providing more reliable input data for subsequent secondary crash prediction models.

### 3.1.2 Predictive Component: Transformer

The predictive component of the VarFusiGAN-Transformer model leverages Transformer architecture to predict both the occurrence probability and spatiotemporal distribution of secondary crashes. The model structure includes an input layer, a Transformer layer, a normalization layer, a dropout layer, a flattening layer, a concatenation layer, a hidden layer, and both classification and regression output layers.

The input layer of the model consists of two parts: dynamic feature input and static feature input. Dynamic features are the dynamic data in the samples. These features are reshaped into a three-dimensional input tensor with the shape (number of samples, number of time steps, number of features per time step). Static features, such as crash type and severity, are discrete data directly used as two-dimensional inputs with the shape (number of samples, number of features).

The output layer of the model consists of two parts that are trained using two loss functions. Among them, the classification output is the binary cross entropy, which is used to predict the occurrence of secondary crashes. The sigmoid activation function maps the output values to the range [0, 1], which indicates the probability of occurrence of secondary crashes. The regression output is the mean square error, and the linear activation function is used to predict the time and distance difference between the primary and secondary crashes.

### 3.2 Evaluation Metrics

The imbalanced real data samples were randomly divided into a training set and a test set in a 7:3 ratio. The sample data generated by different models were then added to the training set to form balanced synthetic data.

To evaluate the accuracy of the classification for the real imbalanced data and the balanced synthetic data, we used three performance metrics: sensitivity, specificity, and G-mean. Sensitivity measures the model's ability to correctly identify positive events, aligning with our primary goal of predicting the minority secondary crashes. It represents the proportion of correctly classified secondary crashes out of all actual secondary crashes. Specificity reflects the model's ability to accurately identify negative events, representing the proportion of correctly predicted primary crashes out of all observed primary crashes. G-mean is a balanced metric that considers the classification accuracy of both positive and negative classes. This metric aims to find a balance between the accuracy of both classes, making it particularly suitable for handling imbalanced datasets. The formulas defining these metrics are:

$$Sensitivity = \frac{TP}{TP + FN} \qquad (4)$$

$$Specificity = \frac{TN}{TN + FP} \qquad (5)$$

$$G-mean = \sqrt{Sensitivity \cdot Specificity} \qquad (6)$$

Additionally, to determine the accuracy of predicting the time difference and distance difference between secondary and primary crashes, we used Mean Absolute Error (MAE) and Root Mean Square Error (RMSE) for evaluation. MAE provides an intuitive error measure by calculating the average absolute error between the predicted values and the actual values, reflecting the overall bias of the prediction. RMSE, on the other hand, calculates the average of the squared differences between the predicted values and the actual values and then takes the square root, making it more sensitive to larger errors in the prediction process. Using these two metrics together allows for a comprehensive evaluation of the impact of different datasets on prediction performance, thereby demonstrating which model-generated data has higher fidelity. Equation (6) and (7) represent the calculation of MAE and RMSE.

$$MAE = \frac{1}{N} \sum_{i=1}^{n} \left| y_i - \hat{y}_i \right| \qquad (6)$$

$$RMSE = \sqrt{\frac{1}{N} \sum_{i=1}^{n} (y_i - \hat{y}_i)^2} \qquad (7)$$

where $y_i$ refers to the real value and $\hat{y}_i$ refers to the predicted value. $N$ is the total number of samples.

# 4 DATA PREPARATION

## 4.1 Data Sources

The objective of this study is to predict the spatiotemporal locations of secondary crashes in real-time based on the characteristics of highway traffic crashes. Therefore, a substantial amount of static crash characteristic data (such as crash type, road conditions, and the spatial and temporal distance between primary and secondary crashes) and dynamic traffic flow characteristic data (such as flow, speed, and occupancy) is required for the training and testing of the model.

This study selected segments of four interstate highways in the United States: I-5, I-90, I-405, and I-520. These highways are equipped with loop vehicle detectors capable of collecting instantaneous traffic flow data, such as volume, occupancy, and speed. In addition, crash datasets from these four interstates were obtained for the period from 2021 to February 2024, totaling 14,409 crashes. This dataset provides detailed records of each crash, including the time and location, type, severity, lighting conditions, and road conditions. The spatiotemporal location of each crash was matched with nearby loop detector data, and unreasonable data were removed. Ultimately, 9,295 crash data samples that met the research requirements were selected.

## 4.2 Identification of Secondary Crashes

According to the introduction in **Section 2**, methods for identifying secondary crashes are generally categorized into static and dynamic approaches. This study employs the speed contour method to identify secondary crashes.

First, a fixed temporal and spatial range is selected to preliminarily screen potential secondary crashes. Based on previous research, the temporal range is typically from 0.25 to 2 hours after the primary crash, and the spatial range is from 1 to 3 kilometers upstream of the primary crash(*18*). Considering that on busy interstate highways, congestion and other subsequent impacts caused by traffic crashes often last longer and affect a broader area, we adopt the maximum ranges used in previous research, namely a 3-mile distance interval and a 2-hour time interval, as the initial screening thresholds.

Next, speed contour plots are generated for the preliminarily screened crashes. Specifically, speed data within the selected time window and spatial range are aggregated at 5-minute intervals to create the initial speed contour plot. To eliminate the interference of recurrent congestion, the average speed on all non-crash days within the same spatiotemporal range is selected, and the difference between these data and the speeds on the day of the crash is calculated to more accurately determine the impact range of the crash on traffic flow.

For example, a crash occurred at mile marker 161.86 at 14:25 on April 17, 2021. Speed data for the stretch from mile markers 156 to 166 between 14:00 and 16:00 on the same day were first extracted, and an initial speed contour plot was generated, as shown in **Figure 5(a)**. After determining the initial impact range of the crash, another crash occurring at mile marker 161.2 at 14:50 on the same day fell within the impact range of the primary crash. The average speed on all non-crash days within this spatiotemporal range was then screened, and the difference between these data and the speeds on the day of the crash was calculated, resulting in a new speed contour plot, as shown in **Figure 5(b)**. The secondary crash remained within the impact range, confirming it as a secondary crash within the impact range of the primary crash.

Using the above method, the impact range of all primary crashes is determined, and any crash occurring within this impact range is identified as a secondary crash. Finally, 9,220 ordinary crashes, 75 primary crashes, and 156 secondary crashes were ultimately identified, with a secondary crash rate of approximately 1.65% being determined. This finding is consistent with existing research, such as 1.2% (Xu et al., 2016), 1.6% (Li and Abdel-Aty, 2022), and 1.98% (Li et al., 2023).

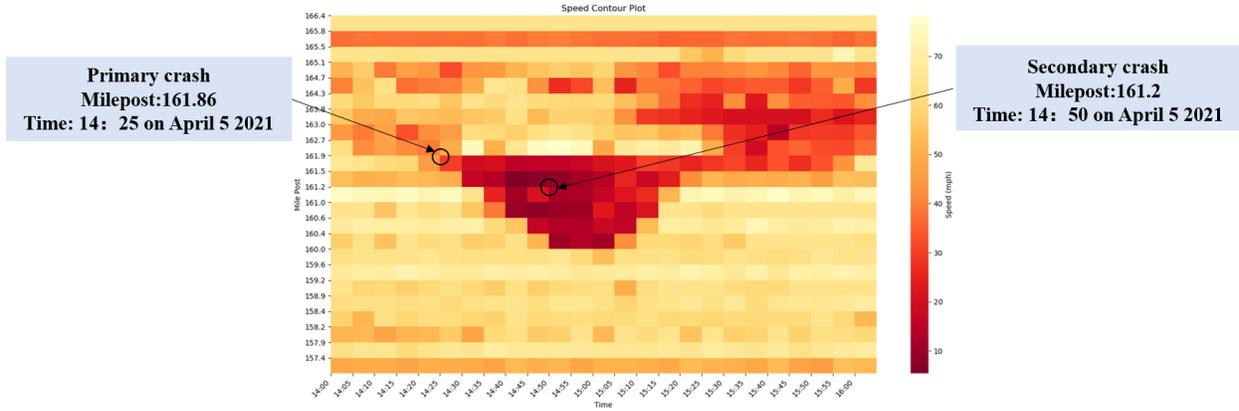

(a) Speed Contour Plot Without Accounting for the Recurrent Congestions

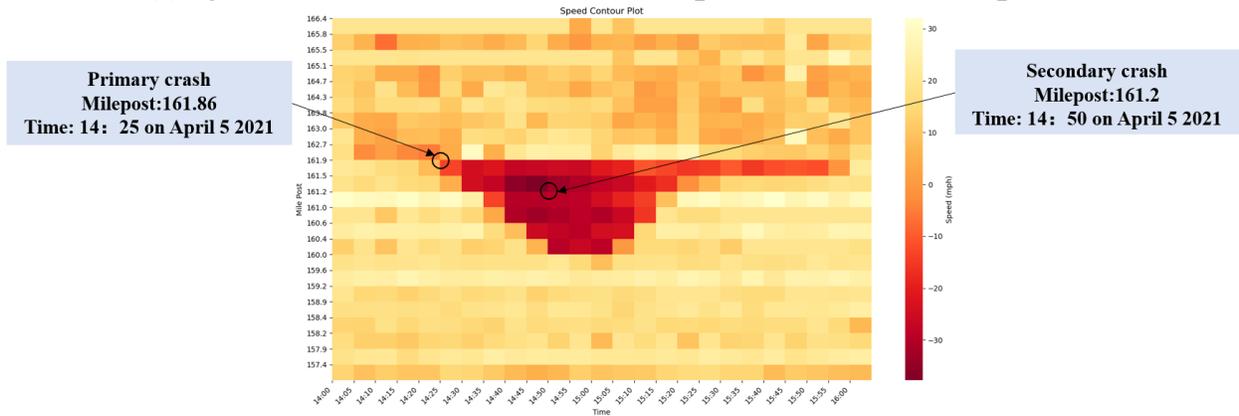

(b) Speed Contour Plot Accounting for the Recurrent Congestions

**Figure 5 An Example of Secondary Crash Identification**

### 4.3 Feature Selection

This study aims to predict the spatiotemporal locations of secondary crashes by combining static crash characteristic data with dynamic traffic flow data. Therefore, in terms of static data, four discrete variables were selected: crash type, crash severity, lighting conditions and road surface conditions. Regarding dynamic data, this study selected traffic flow data collected from one loop detector upstream and one downstream of the primary crash location within the 30 minutes preceding the crash, with a time interval of 5 minutes. **Table 1** provides the statistics and descriptions of all variables. For example, "Up_Avg_Occ" denotes the average occupancy upstream every 5 minutes, "Down_Dif_AvgFlow" denotes the absolute value of the average flow difference in adjacent lanes downstream every 5 minutes, and "Updown_AvgSpd" denotes the average speed difference between upstream and downstream every 5 minutes.

### 4.4 Experimental Design

To comprehensively evaluate the predictive performance of the VarFusiGAN-Transformer model, this study first assesses the fidelity of the data generated by VarFusiGAN. For the evaluation of the fidelity of this generative model, we selected the following four models for comparative evaluation: MC-GAN(*38*), TVAE(*39*), TimeGAN(*7*), and TimeVAE(*9*). Among them, MC-GAN can generate multidimensional complex data under multiple conditional constraints, and TVAE is specifically designed to handle mixed-type tabular data. Therefore, we chose these two models to compare the ability of VarFusiGAN to generate composite data. In addition, dynamic data often play a crucial role in traffic crash data. Thus, we chose TimeGAN and TimeVAE to compare and evaluate VarFusiGAN's performance in generating dynamic data. However, since TimeGAN and TimeVAE cannot generate composite data, the data they generate are unsuitable for evaluating static data and cannot be used for subsequent secondary crash prediction.

This study used resampling to balance the real training dataset. The training and testing datasets include both the real dataset and the balanced dataset. Seventy percent of the crash data (109 secondary crashes and 6,454 primary crashes) were randomly selected for training, while the remaining 30% (47 secondary crashes and 2,766 primary crashes) were used for testing. For the resampled balanced training dataset, the ratio of secondary crashes to primary crashes was 1:1. Therefore, 6,345 secondary crash samples were generated using VarFusiGAN, MC-GAN, and TVAE. Additionally, to verify which ratio of the training dataset produces better prediction accuracy, we constructed training datasets with secondary to primary crash ratios of 1:4, 1:3, and 1:2 using secondary crash samples generated by VarFusiGAN.

Since the four comparative models mentioned above cannot handle samples of different lengths, it is not possible to directly input complete secondary crash samples into these models. Therefore, the experiments were divided into two steps. In the first step, for all secondary crash samples, the dynamic features after the primary crash were removed, retaining only all static data and dynamic data from the 30 minutes before the primary crash to ensure that all samples have the same length. The trimmed dataset was then input into VarFusiGAN and the other four comparative models to evaluate the fidelity of the generated data. Subsequently, the balanced training sets generated by the five models were input into the prediction model to verify the improvement in prediction results compared to the real dataset. In the second step, the complete untrimmed dataset was input into the VarFusiGAN-Transformer model, and the results of the generation and prediction experiments were compared with those from the first step to demonstrate that the VarFusiGAN-Transformer model can achieve high-quality generation and accurate prediction of secondary crash data.

**TABLE 1 Summary of Variables Statistical Description (Primary Crashes)**

| Variables (static) | Description | Count | Secondary Crash | Normal Crash |
|---|---|---|---|---|
| Crash type | Real-end = 1 | 6048 | 121 | 5927 |
| | Sideswipe/Angle/Head-On Crash = 2 | 2946 | 33 | 2913 |
| | Obstacle Crash = 3 | 187 | 2 | 185 |
| | Other non-crash = 4 | 195 | 0 | 195 |
| Crash severity | Minor = 1 | 7043 | 121 | 6922 |
| | Moderate = 2 | 2218 | 33 | 2185 |
| | Severe = 3 | 115 | 2 | 113 |
| Lighting conditions | Bright = 1 | 7114 | 125 | 6989 |
| | Transitional = 2 | 1788 | 18 | 1770 |
| | Dark = 3 | 474 | 13 | 461 |
| Road surface conditions | Dry = 1 | 7363 | 109 | 7254 |
| | Wet = 2 | 1833 | 43 | 1790 |
| | Other = 3 | 180 | 4 | 176 |

| Variables (dynamic) | Secondary Crash | | | | Normal Crash | | | |
|---|---|---|---|---|---|---|---|---|
| | Mean | Std | Max | Min | Mean | Std | Max | Min |
| Up_Avg_Flow | 78.26 | 32.14 | 159.17 | 0.33 | 77.28 | 32.91 | 211.33 | 0.17 |
| Up_Avg_Occ | 13.50 | 9.09 | 45.75 | 1.75 | 13.46 | 10.56 | 77.63 | 0.04 |
| Up_Avg_Spd | 46.46 | 18.59 | 74.83 | 4.40 | 46.63 | 18.10 | 82.29 | 0.50 |
| Up_Dif_AvgFlow | 6.28 | 4.92 | 29.08 | 1.42 | 5.18 | 3.63 | 29.11 | 0.08 |
| Up_Dif_AvgOcc | 9.77 | 10.75 | 55.25 | 0.61 | 9.57 | 9.89 | 77.00 | 0.06 |
| Up_Dif_AvgSpd | 6.66 | 7.96 | 41.58 | 0.67 | 5.97 | 6.47 | 77.33 | 0.06 |
| Down_Avg_Flow | 79.55 | 30.42 | 142.50 | 4.67 | 71.85 | 31.90 | 223.67 | 0.17 |
| Down_Avg_Occ | 15.88 | 10.63 | 50.54 | 1.17 | 13.52 | 10.95 | 77.08 | 0.03 |
| Down_Avg_Spd | 6.73 | 5.92 | 32.08 | 0.33 | 5.87 | 6.43 | 59.72 | 0.06 |
| Down_Dif_AvgFlow | 46.33 | 18.48 | 82.13 | 9.29 | 46.34 | 18.82 | 86.37 | 0.37 |
| Down_Dif_AvgOcc | 5.55 | 3.86 | 21.83 | 0.83 | 5.05 | 3.57 | 32.50 | 0.06 |
| Down_Dif_AvgSpd | 9.72 | 9.34 | 42.50 | 0.75 | 9.70 | 10.38 | 78.83 | 0.06 |
| Updown_AvgFlow | 26.93 | 20.04 | 100.67 | 5.17 | 22.69 | 17.26 | 154.67 | 0.17 |
| Updown_AvgOcc | 7.70 | 5.84 | 25.93 | 0.46 | 6.25 | 6.03 | 74.53 | 0.04 |
| Updown_AvgSpd | 14.07 | 12.54 | 62.87 | 1.73 | 14.33 | 12.76 | 77.00 | 0.03 |

Note:

Up_Avg_Flow: Average vehicle flow during 5-min period upstream (veh/30 s)

Up_Avg_Occ: Average detector occupancy during 5-min period upstream (%)

Up_Avg_Spd: Average vehicle speed during 5-min period upstream (mph)

Up_Dif_AvgFlow: Average vehicle flow difference between adjacent lanes upstream (veh/30 s)

Up_Dif_AvgOcc: Average detector occupancy difference between adjacent lanes upstream (%)

Up_Dif_AvgSpd: Average vehicle speed difference between adjacent lanes upstream (mph)

Down_Avg_Flow: Average vehicle flow during 5-min period downstream (veh/30 s)

Down_Avg_Occ: Average detector occupancy during 5-min period downstream (%)

Down_Avg_Spd: Average vehicle speed during 5-min period downstream (mph)

Down_Dif_AvgFlow: Average vehicle flow difference between adjacent lanes downstream (veh/30 s)

Down_Dif_AvgOcc: Average detector occupancy difference between adjacent lanes downstream (%)

Down_Dif_AvgSpd: Average vehicle speed difference between adjacent lanes downstream (mph)

Updown_AvgFlow: Average vehicle flow difference between upstream and downstream (veh/30 s)

Updown_AvgOcc: Average detector occupancy difference between upstream and downstream (%)

Updown_AvgSpd: Average vehicle speed difference between upstream and downstream (mph)

**5 RESULTS**

**5.1 Evaluation of Generated Data**

In this section, we adopt various evaluation methods to comprehensively analyze the performance of the VarFusiGAN model on dynamic and static data. First, in Section 5.1.1, we extracted the maximum and minimum values of dynamic variables at each time step and calculated their average to analyze the distribution of dynamic features. At the same time, the distribution of discrete variables is also presented. Next, in Section 5.1.2, we calculated the joint distribution density heatmap between two variables and compared the distribution characteristics of real data and generated data to assess the model's ability to reproduce the real data distribution. In Section 5.1.3, we calculated the Pearson correlation coefficient distribution between dynamic variables to evaluate the model's performance in capturing relationships between variables. The above three sections evaluate the fidelity of the trimmed generated data. Finally, in Section 5.1.4, we conducted the same fidelity evaluation for three datasets: untrimmed real data, trimmed VarFusiGAN-generated data, and untrimmed VarFusiGAN-generated data, to comprehensively compare the fidelity of different datasets.

*5.1.1 Conditional Distribution*

To reflect the variation characteristics of dynamic data over the entire time period, we extracted the maximum and minimum values of each dynamic variable at all time steps and calculated their means. Subsequently, we calculated the distribution of these means for each dynamic feature. This distribution avoids the influence of random factors or noise and can capture the impact of extreme values on the entire time series. **Figures 6(1)-(6)** show the distribution histograms of the "Updown_AvgSpd" variable. The horizontal coordinates represent the possible values and the vertical coordinates are the density values of the variable. It can be observed that the DG model can effectively learn the characteristics of dynamic data, especially capturing the long-tail distribution excellently. Even for time series data of different lengths, it can have more desirable distribution characteristics. In comparison, the distributions of data generated based on MC-GAN and TimeVAE show much greater deviations. Although TimeGAN and TVAE have also learned some of the tail features to some extent, they exhibit peak shift problems. Similarly, for discrete variables of static data, the DG model also performs well (**Figures 6(7)-(10)**). For example, when considering crash severity, MC-GAN did not generate data with the correct proportional distribution, while TVAE experienced mode collapse, with the minority class "Severe" almost disappearing.

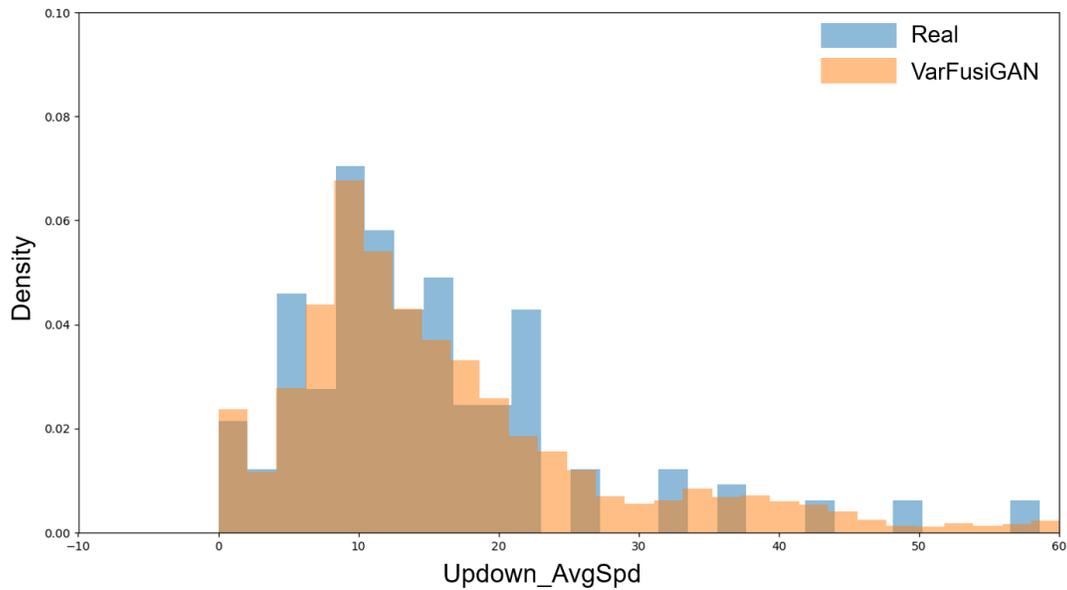

(1) Updown_AvgSpd (VarFusiGAN)

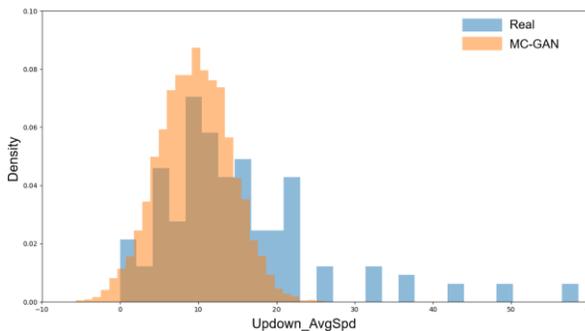

(2) Updown_AvgSpd (MC-GAN)

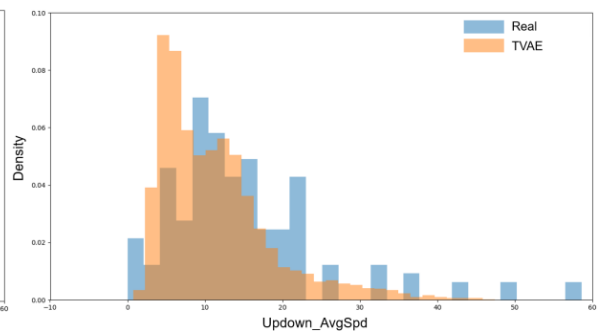

(3) Updown_AvgSpd (TVAE)

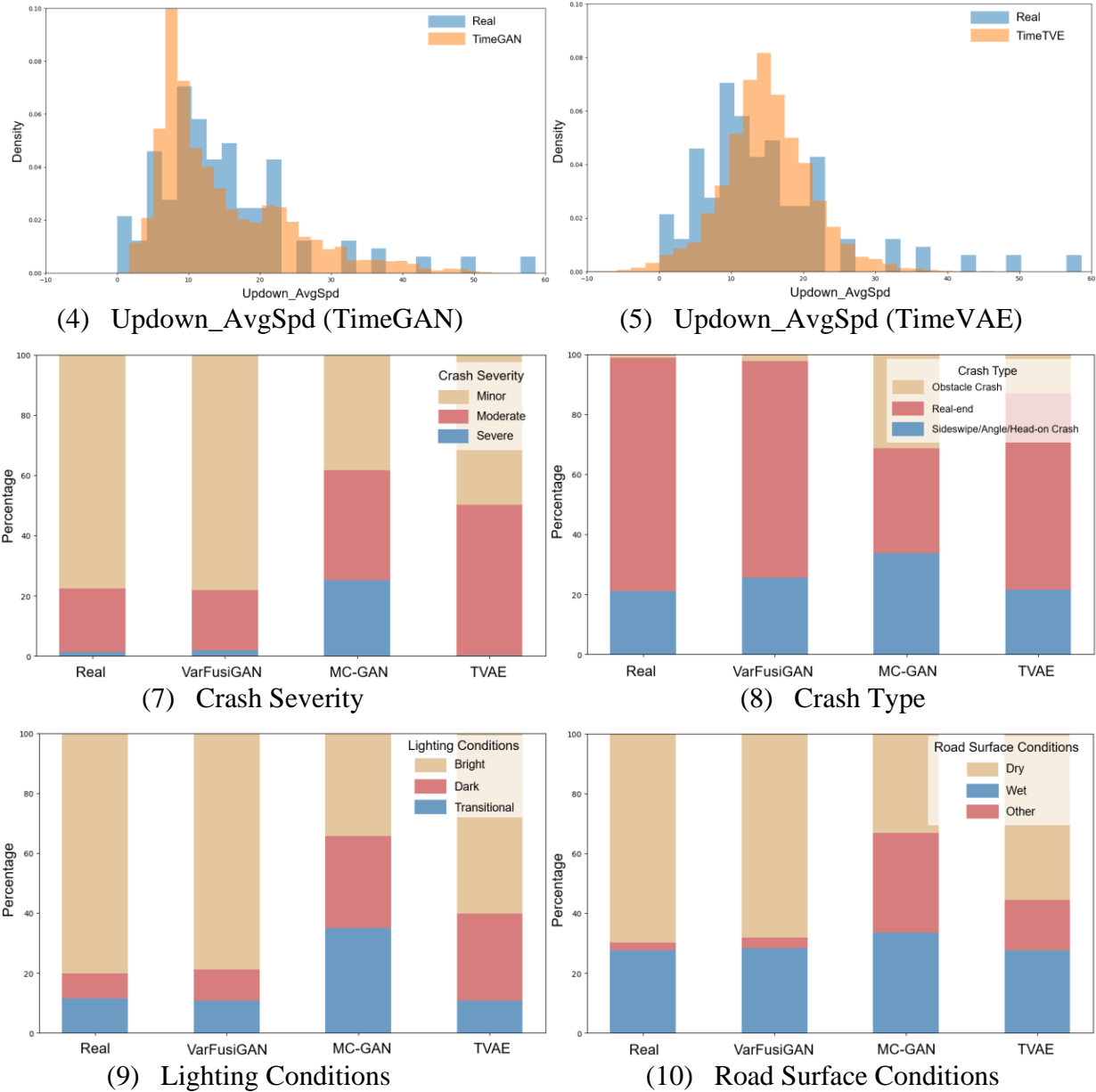

(4) Updown_AvgSpd (TimeGAN)  (5) Updown_AvgSpd (TimeVAE)

(7) Crash Severity  (8) Crash Type

(9) Lighting Conditions  (10) Road Surface Conditions

**Figure 6 Conditional Distribution of Variables**

### 5.1.2 Joint Distribution

Additionally, to further compare the distribution characteristics of real and generated data, we calculated the joint distribution density heatmap for variables such as Crash type, Road surface conditions, Updown_AvgFlow, and Down_Dif_AvgFlow. The darker the color, the higher the density. As shown in **Figures 7 (1-10)**, the joint distribution deviations of the data generated by DG are much smaller than those generated by MC-GAN, TVAE, TimeGAN, and TimeVAE.

**Figures 7 (2) and (8)** indicate that the high-density areas presented by the data generated by DG are consistent with the distribution of real data (**Figures 7 (1) and (7)**), even for low-probability events. In contrast, the joint distributions generated by the other models generally exhibit various flaws, including differences in the number, shape, and location of density areas.

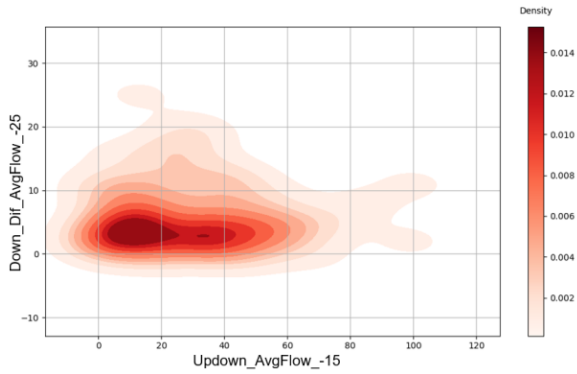

(1)   Real (dynamic)

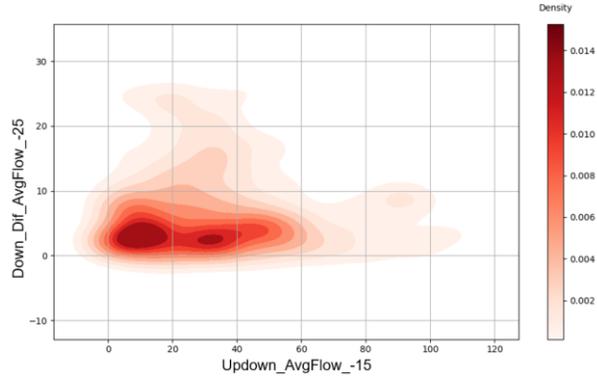

(2)   DoppelGANger (dynamic)

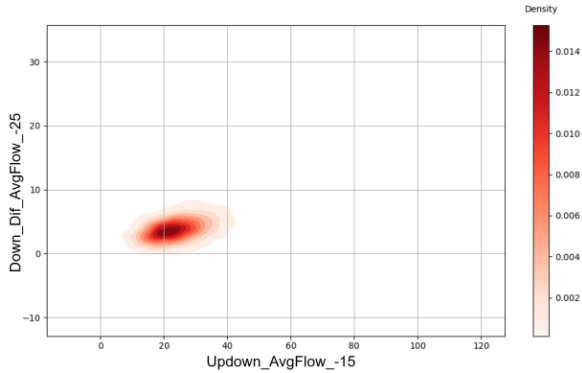

(3)   MC-GAN (dynamic)

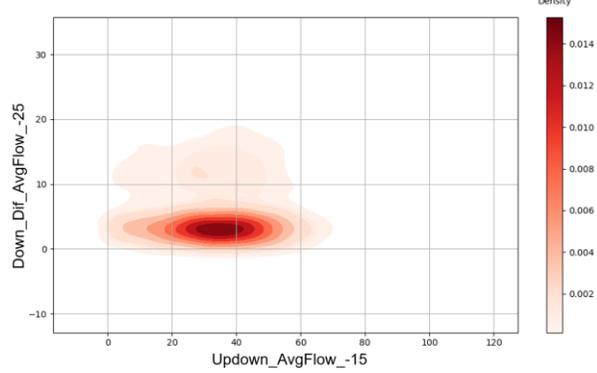

(4)   TVAE (dynamic)

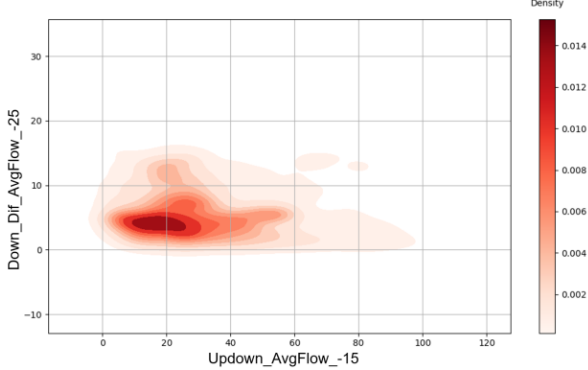

(5)   TimeGAN (dynamic)

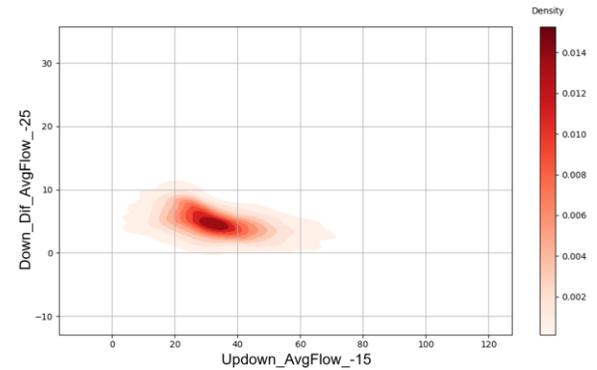

(6)   TimeVAE (dynamic)

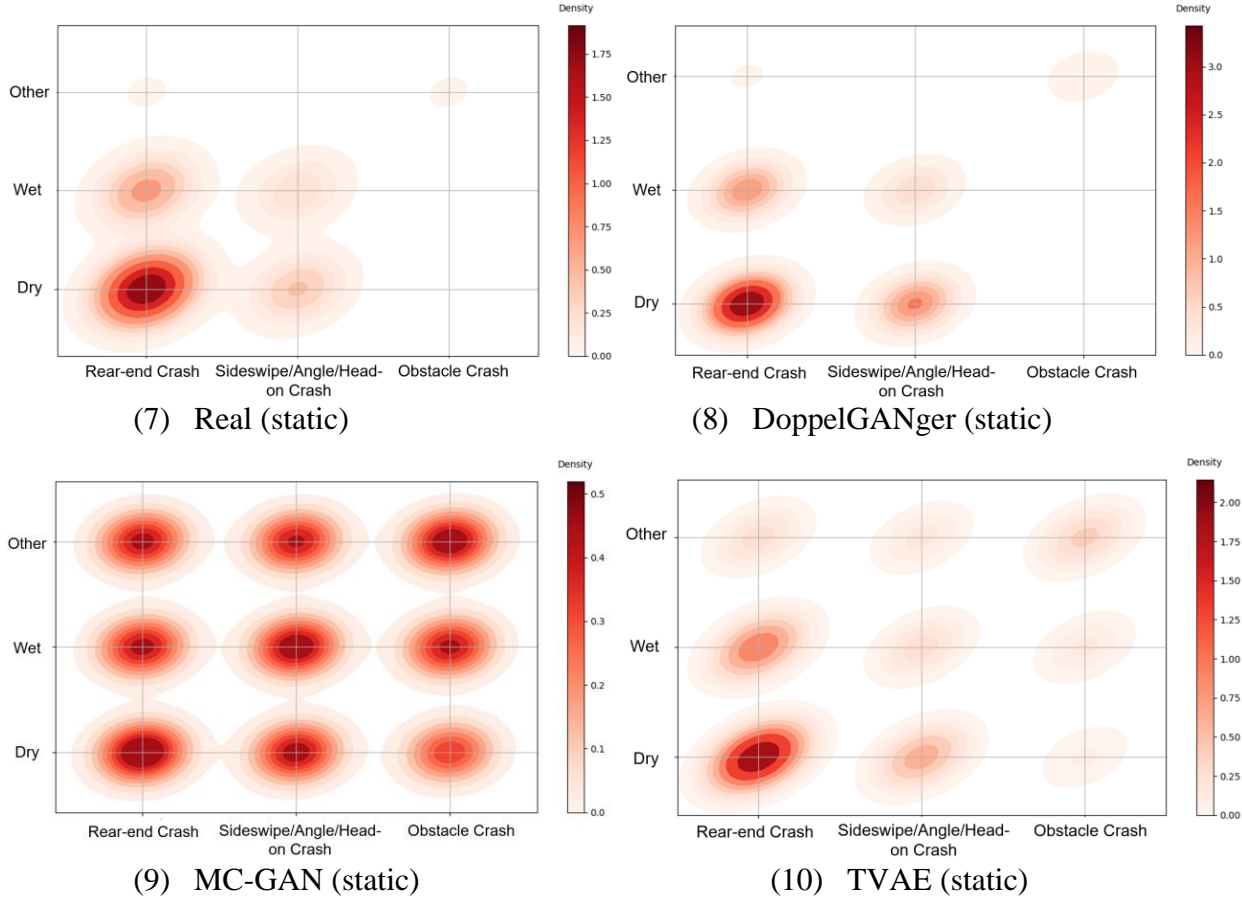

**Figure 7 Joint Distribution of Variables**

### 5.1.3 Correlation Coefficient Distribution

Figures 8 shows the distribution of Pearson correlation coefficients for some dynamic variables across the entire time series. The horizontal coordinates represent the 6-time steps for each dynamic variable, and the vertical coordinates represent the range of Pearson correlation coefficients between those two variables. We observe that VarFusiGAN performs better in capturing the correlations between variables, whereas the performance of other models is relatively poor. This demonstrates that VarFusiGAN can not only generate high-fidelity univariate time series data but also reproduce the complex relationships between different dynamic variables, which is crucial for subsequent predictive tasks that rely on inter-variable relationships.

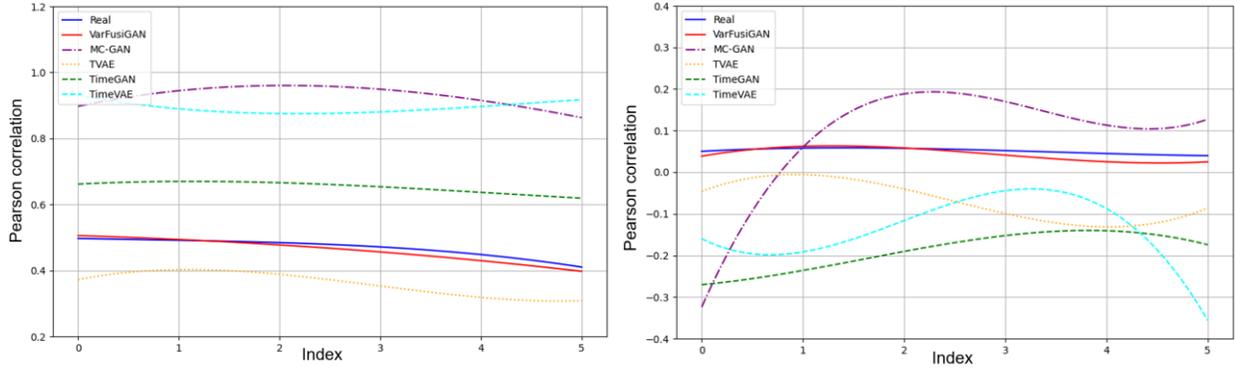

**Figure 8 Pearson Correlation Coefficient Distribution of Dynamic Variables**

*5.1.4 Evaluation of Generated Data for Sample Data of Different Length*

To demonstrate that the VarFusiGAN model maintains high fidelity when generating untrimmed data, this section conducts the same fidelity evaluation for three datasets: untrimmed real data, trimmed VarFusiGAN-generated data, and untrimmed VarFusiGAN-generated data..

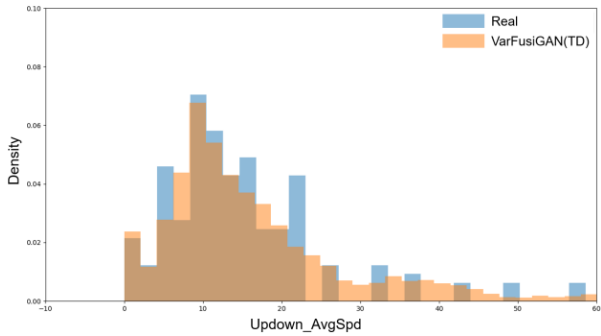

(1) Updown_AvgSpd (VarFusiGAN TD)

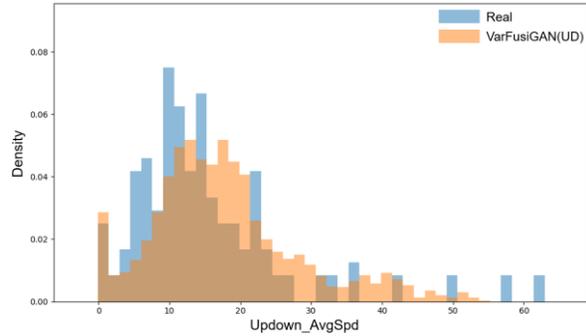

(2) Updown_AvgSpd (VarFusiGAN UD)

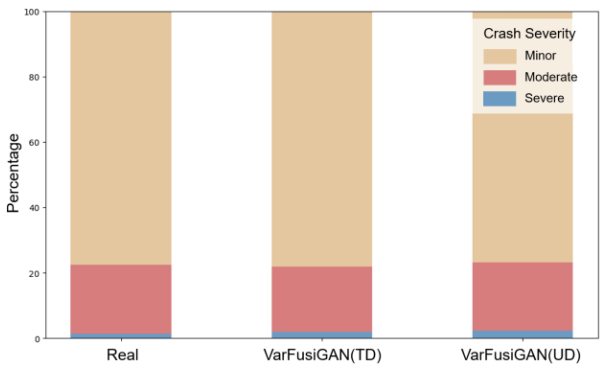

(3) Crash severity

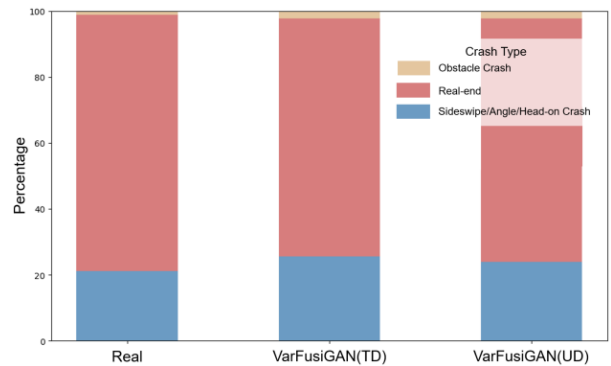

(4) Crash type

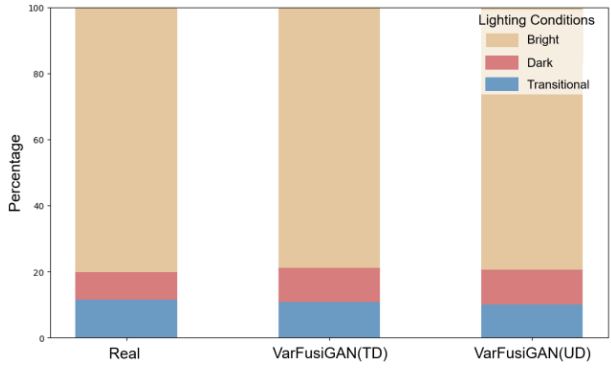
(5)   Lighting conditions

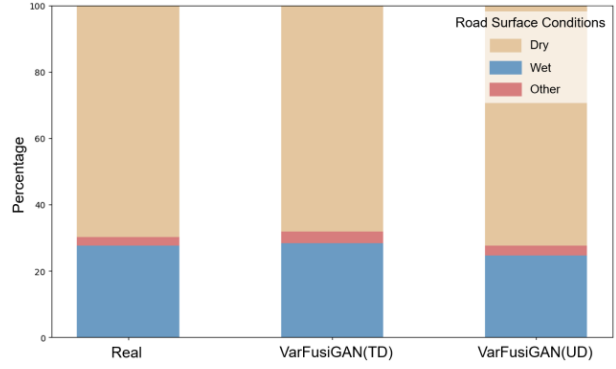
(6)   Road surface conditions

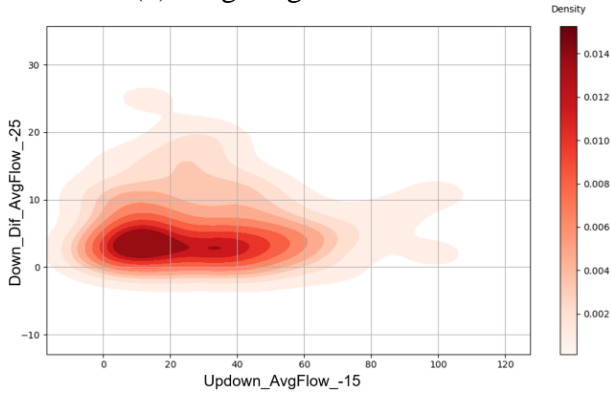
(7)   Real (dynamic)

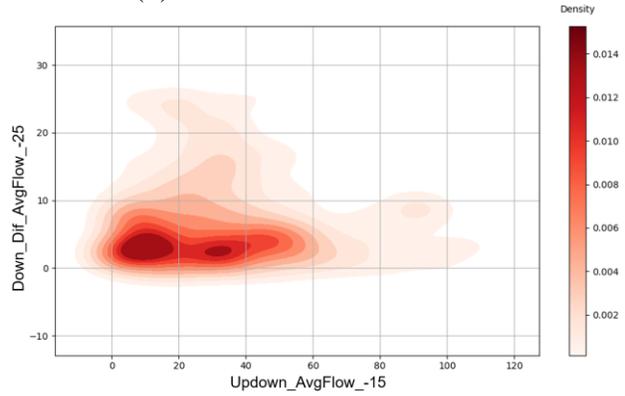
(8)   VarFusiGAN UD (dynamic)

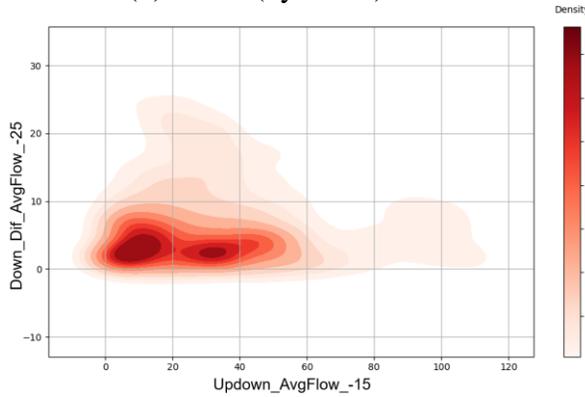
(9)   VarFusiGAN TD (dynamic)

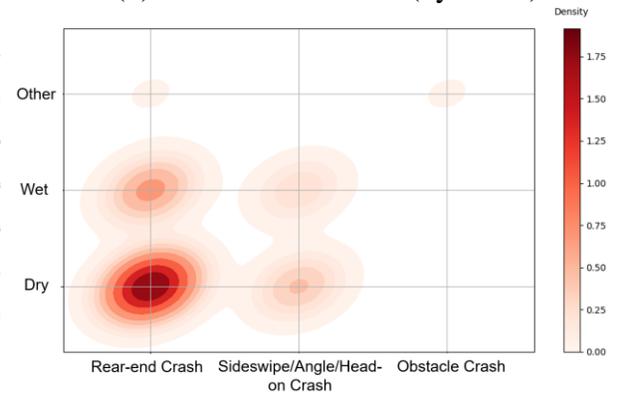
(10)   Real (static)

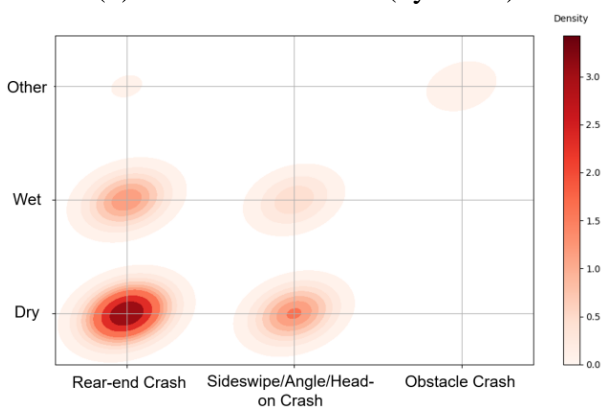
(11)   VarFusiGAN UD (static)

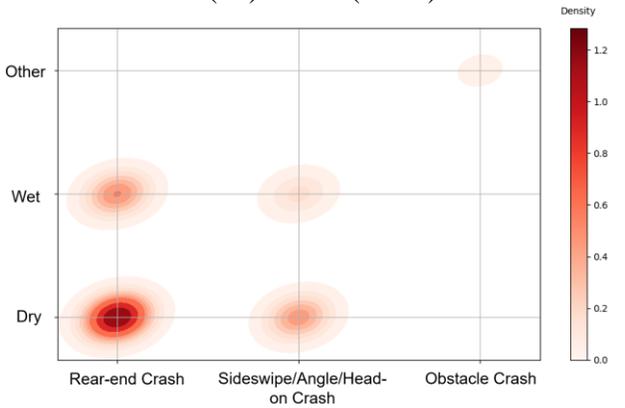
(12)   VarFusiGAN TD (static)

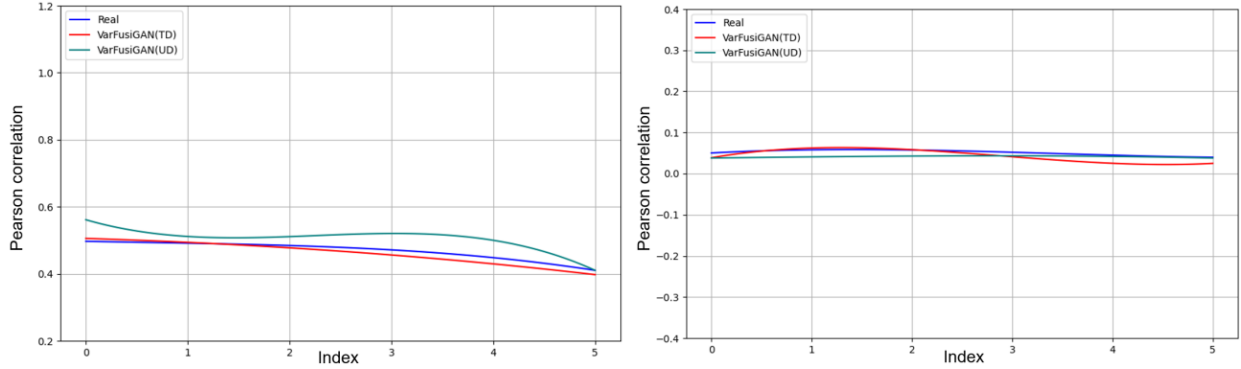

(13)  Pearson Correlation Coefficient Distribution of Dynamic Variables

**Figure 11 Results of the Assessment of Each Characterization Indicator for the Three Types of Datasets**

Note:

VarFusiGAN UD: Untrimmed data generated by VarFusiGAN

VarFusiGAN TD: Trimmed data generated by VarFusiGAN

As shown in **Figure 9**, the VarFusiGAN model effectively generates samples with different lengths (i.e., untrimmed) of dynamic data while maintaining a high level of fidelity across various features. For dynamic features such as "Updown_AvgSpd," both types of generated data from the VarFusiGAN model show a distribution closely aligned with the real data, even capturing subtle fluctuations and extreme values. This consistency indicates that the VarFusiGAN model can handle the complexity introduced by different time series lengths, generating realistic data samples regardless of the time span. Regarding static features such as "crash severity," "crash type," "lighting conditions," and "road surface conditions," the VarFusiGAN model demonstrates an impressive ability to reproduce the proportional distribution of each feature. The generated data closely matches the real data's distribution, further proving the model's effectiveness in preserving the underlying relationships between static attributes, irrespective of time series length. The Pearson correlation coefficient distribution shown in **Figure 9 (13)** also indicates that the VarFusiGAN model performs excellently in capturing inter-variable correlations, regardless of the dynamic data length. This demonstrates that generating samples of different lengths does not compromise the model's ability to maintain dynamic relationships between variables. Overall, the VarFusiGAN model achieves high fidelity in both dynamic and static variables, and the generated data closely aligns with the real data, showcasing the robustness of the VarFusiGAN model in handling complex datasets with different lengths.

## 5.2 Evaluation of Prediction Accuracy

### 5.2.1 Evaluation of regression and classification results

This section evaluates the prediction results of balanced datasets based on trimmed and untrimmed data, both at a 1:1 ratio. As described in **Section 4.4**, the prediction of secondary crashes in this study involves two steps: classification and regression. The evaluation metrics for classification include sensitivity, specificity, and G-mean, while for regression, the metrics are Mean Absolute Error (MAE) and Root Mean Square Error (RMSE). The results are shown in **Table 2**.

In terms of classification, it is evident that the prediction performance based on the original data is extremely poor, regardless of time or distance, as the model fails to identify secondary crashes, which constitute the minority class. In contrast, the Transformer model trained on balanced datasets consistently exhibits higher sensitivity and G-mean, albeit with a decrease in specificity. Among all the models, the VarFusiGAN-Transformer model achieved the best classification performance, with G-mean values of 0.966 and 0.970, respectively, indicating that the data generated by the VarFusiGAN model provides higher accuracy in determining the occurrence of secondary crashes. Moreover, the untrimmed data, when trained with the VarFusiGAN-Transformer, yielded the highest sensitivity and G-mean values. This phenomenon aligns with our earlier statement that a more complete data structure helps the model extract more internal features, thereby improving classification accuracy.

In terms of regression, the balanced datasets also led to a noticeable improvement in the prediction accuracy for both time and distance. Specifically, while the data generated by the MC-GAN model improved prediction performance to some extent, its error values remained relatively high, particularly for spatial predictions, where it underperformed compared to the other two models. The TVAE model performed slightly better, with prediction errors significantly lower than those of the original dataset, though still insufficient for meeting the prediction requirements. Undoubtedly, the balanced dataset based on the VarFusiGAN model performed the best, with the lowest MAE and RMSE values for both time and distance predictions, indicating that the generated data substantially enhances regression prediction accuracy. Similar to the classification results, when using untrimmed data as input, the VarFusiGAN-Transformer model achieved the best prediction results, demonstrating that complete datasets can best preserve the spatiotemporal relationships between static and dynamic features, thereby allowing the prediction model to achieve more optimal predictive performance.

**TABLE 2 Prediction Results Based on Different Models**

| | Measurements | | Real Dataset | VFGAN (TD) | VFGAN (UD) | MC-GAN | TVAE |
|---|---|---|---|---|---|---|---|
| Classification | Sensitivity | | 0.000 | 0.972 | 0.973 | 0.662 | 0.989 |
| | Specificity | | 1.000 | 0.959 | 0.968 | 0.953 | 0.903 |
| | G-mean | | 0.000 | 0.966 | 0.970 | 0.794 | 0.945 |
| Regression | Time Difference | MAE | 0.867 | 0.355 | 0.349 | 0.680 | 0.483 |
| | | RMSE | 1.043 | 0.458 | 0.448 | 0.807 | 0.663 |
| | Distance Difference | MAE | 0.885 | 0.495 | 0.469 | 0.764 | 0.521 |
| | | RMSE | 1.167 | 0.640 | 0.591 | 1.014 | 0.671 |

Note:
VFGAN(UD): Untrimmed data generated by VarFusiGAN
VFGAN(TD): Trimmed data generated by VarFusiGAN

*5.2.2 Joint Evaluation of Spatiotemporal Prediction*

Although MAE and RMSE provide a quantitative evaluation to compare the quality of different datasets, they do not intuitively reflect the actual prediction accuracy. Therefore, to more clearly demonstrate the achievable prediction accuracy of different datasets in practical scenarios, we conducted a joint temporal and spatial evaluation for the three balanced datasets based on untrimmed data.

First, all correctly classified secondary crash samples in the test set were selected. Using the true values of each sample as the origin, a Cartesian coordinate plane was plotted with the x-axis representing time and the y-axis representing distance, defining a fixed time-distance range on the graph. Next, the absolute difference between the predicted and true values for each sample was calculated, and the point with this difference as its coordinate was plotted as the predicted value. Finally, the prediction accuracy was determined based on whether the predicted value fell within the aforementioned fixed range. For each dataset, the ratio of the number of samples within the range to those outside the range represents its prediction accuracy. Therefore, as the range narrows, the prediction accuracy gradually decreases.

In this study, we created three different time-distance ranges: 1 hour × 1 mile, 0.5 hours × 0.5 miles, and 0.2 hours × 0.2 miles, to eliminate chance occurrences and comprehensively evaluate the prediction performance of the datasets. Figure 10 shows the accuracy within the aforementioned ranges for balanced datasets constructed from the data generated by VarFusiGAN, MC-GAN, and TVAE. It can be observed that, regardless of the range, the synthetic data based on VarFusiGAN consistently exhibits the highest accuracy. Moreover, as the time-distance range narrows, the prediction accuracy of the dataset based on VarFusiGAN shows a more significant advantage compared to the datasets from the other two models. Overall, VarFusiGAN achieved better performance in the generated data performance evaluation, further demonstrating that the VarFusiGAN-Transformer model can effectively predict the temporal and spatial occurrence of secondary crashes.

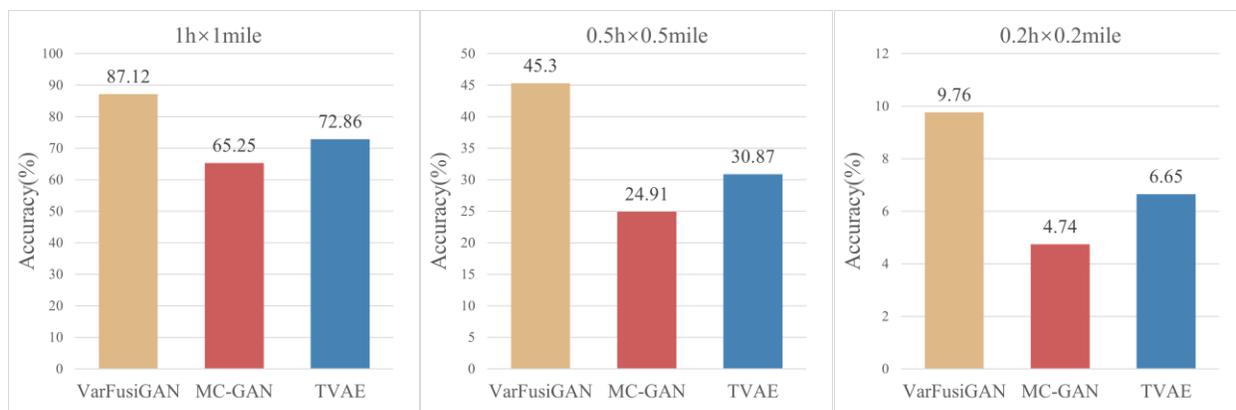

(a) 1 hour × 1 mile   (b) 0.5 hour × 0.5 mile   (c) 0.2 hour × 0.2 mile

**Figure 10 Joint Evaluation of Spatiotemporal Prediction with Different Ranges**

### 5.3 Predictive effects of different scaled balanced data

In this section, the trimmed (equal-length) data and untrimmed (different-length) data are respectively fed into the VarFusiGAN model for generation, and the latter was composed into multiple balanced datasets with different ratios (1:4, 1:3, 1:2, 1:1) to determine which ratio produced the best prediction accuracy. The summarized results in Table 4 indicate that, for both classification and regression tasks, the prediction accuracy of balanced datasets composed of untrimmed data improved as the balance ratio increased, with the best performance achieved at a balance ratio of 1:1, outperforming the datasets composed of trimmed data.

In the classification task, all datasets consistently showed improvements compared to the real data under different balance ratios (1:4, 1:3, 1:2, 1:1). At a balance ratio of 1:1, the sensitivity (0.973) and G-mean (0.970) based on the untrimmed dataset were the highest. These metrics indicate that more complete dynamic data can help the VarFusiGAN-Transformer model strike a more effective balance between identifying secondary crashes and minimizing false positives.

In the regression task, predictions based on untrimmed data were better than those based on trimmed datasets, suggesting that more detailed dynamic data improves the prediction accuracy of the Transformer model. Specifically, at a balance ratio of 1:1, the MAE and RMSE values of the time difference dropped to 0.349 and 0.448, respectively, indicating that the VarFusiGAN-Transformer performed best at capturing dynamic-static relationships and predicting specific spatiotemporal locations under this ratio.

Overall, the results confirm the robustness of the VarFusiGAN-Transformer model in handling complex datasets with varying lengths. The generated datasets maintained high fidelity and performed excellently in predictive tasks, highlighting the model's versatility in generating data suitable for complex traffic safety analyses and predicting the spatiotemporal location of secondary crashes.

**TABLE 4 Prediction Results Based on Different Balance Ratios**

| | Different Ratio | | Real (≈1:60) | VFG-UT (1:4) | VFG-UD (1:3) | VFG-UD (1:2) | VFG-UD (1:1) | VFG-TD (1:1) |
|---|---|---|---|---|---|---|---|---|
| Classification | Sensitivity | | 0.000 | 0.766 | 0.898 | 0.968 | 0.973 | 0.972 |
| | Specificity | | 1.000 | 0.948 | 0.965 | 0.963 | 0.968 | 0.959 |
| | G-mean | | 0.000 | 0.852 | 0.931 | 0.965 | 0.970 | 0.966 |
| Regression | Time Difference | MAE | 0.710 | 0.532 | 0.415 | 0.411 | 0.349 | 0.355 |
| | | RMSE | 0.897 | 0.647 | 0.501 | 0.502 | 0.448 | 0.458 |
| | Distance Difference | MAE | 0.955 | 0.673 | 0.558 | 0.578 | 0.469 | 0.495 |
| | | RMSE | 1.271 | 0.876 | 0.704 | 0.751 | 0.591 | 0.640 |

Note:
VFG-UD: Untrimmed data generated by VarFusiGAN
VFG-TD: Trimmed data generated by VarFusiGAN

# 6 CONCLUSIONS

In this study, we propose a novel generative adversarial network model called DoppelGANger to predict the spatiotemporal locations of secondary crashes. This model addresses many challenges in generating and predicting complex traffic crash data. By generating multidimensional dynamic and static traffic crash data, DoppelGANger significantly improves the accuracy of predicting the spatiotemporal distribution of secondary crashes. Multiple evaluations demonstrate that the DoppelGANger model performs superiorly in generating multidimensional traffic crash data with high fidelity and outperforms other generative models such as MC-GAN and TVAE. In addition, the experimental results show that the Transformer model trained on the DG-balanced data significantly improves the sensitivity and G-mean of the classification of secondary crashes, and reduces the mean absolute error (MAE) and root mean square error (RMSE) of time and distance prediction. This suggests that the synthetic data generated by DoppelGANger enhances the prediction accuracy of secondary crash occurrence and its spatiotemporal distribution.

In the future, we will optimize the structure of DoppelGANger and consider incorporating a physical system simulator to enhance the model's ability to handle causal interactions between stateful agents. Meanwhile, we note that the training of the current DoppelGANger model requires a large amount of time and computational resources, so we will investigate more efficient optimization algorithms or training techniques in the future to enhance the training efficiency. In addition, many samples with missing variables were directly deleted during the data preparation process in this study. To address this issue, subsequent studies will try to use data filling methods to improve the completeness of the samples.

**ACKNOWLEDGMENTS**

The research is supported by Southeast University Interdisciplinary Research Program for Young Scholars (Grant No. 2024FGC1006), and Shenzhen Technology Program Project (Grant No. SGDX20230821095159012).

**AUTHOR CONTRIBUTIONS**